%% file: aaai25.tex
\documentclass[letterpaper]{article} 
\usepackage{aaai25}  
\usepackage{times}  
\usepackage{helvet}  
\usepackage{courier}  
\usepackage[hyphens]{url}  
\usepackage{graphicx} 
\usepackage{natbib}  
\usepackage{caption} 
\usepackage{comment}
\usepackage{times}
\usepackage{soul}
\usepackage{url}
\usepackage[utf8]{inputenc}
\usepackage{graphicx}
\usepackage{amsmath}
\usepackage{amsfonts}
\usepackage{amsthm}
\usepackage{booktabs}
\usepackage[switch]{lineno}
\usepackage{algorithm}
\usepackage{algorithmicx}
\usepackage{algc}
\usepackage{algcompatible}
\usepackage{algpseudocode}
\usepackage[english]{babel}
\usepackage{underscore}
\usepackage{comment}
\usepackage{cleveref}
\usepackage{subcaption}
\usepackage{enumitem}
\usepackage{lipsum}
\usepackage{dsfont}
\usepackage{comment}
\usepackage{newfloat}
\usepackage{listings}
\DeclareCaptionStyle{ruled}{labelfont=normalfont,labelsep=colon,strut=off} 
\floatstyle{ruled}
\newfloat{listing}{tb}{lst}{}
\floatname{listing}{Listing}
\title{Active Geospatial Search for Efficient Tenant Eviction Outreach}

\author{
    Anindya Sarkar$^1$,
    Alex DiChristofano$^1$$^*$,
    Sanmay Das$^2$,
    Patrick J. Fowler$^1$$^*$$^+$,\\
    Nathan Jacobs$^1$,
    Yevgeniy Vorobeychik$^1$
}
\affiliations{
    \textsuperscript{\rm 1}Washington University in St. Louis, USA;
    \textsuperscript{\rm 2}George Mason University, USA \\

    Department of Computer Science and Engineering\\
    Division of Computational \& Data Sciences$^*$;
    Brown School at Washington University in St. Louis$^+$
    
}

\begin{document}

\maketitle

\begin{abstract}
Tenant evictions threaten housing stability and are a major concern for many cities. An open question concerns whether data-driven methods enhance outreach programs that target at-risk tenants to mitigate their risk of eviction.
We propose a novel \emph{active geospatial search (AGS)} modeling framework for this problem. 
AGS integrates property-level information in a search policy that identifies a sequence of rental units to canvas to both determine their eviction risk and provide support if needed.
We propose a hierarchical reinforcement learning approach to learn a search policy for AGS that scales to large urban areas containing thousands of parcels, balancing exploration and exploitation and accounting for travel costs and a budget constraint.
Crucially, the search policy adapts online to newly discovered information about evictions.
Evaluation using eviction data for a large urban area demonstrates that the proposed framework and algorithmic approach are considerably more effective at sequentially identifying eviction cases than baseline methods.
\end{abstract}

\input{intro}
\input{rel_work}

\input{model}
\input{method}
\input{results}

\input{conclusion}

\section*{Acknowledgements}

This research was partially supported by the National Science Foundation (IIS-1939677, IIS-2214141, IIS-1905558, CNS-2310470, 2402856), Office of Naval Research (N000142412663), Amazon, and NVIDIA.

\bibliography{aaai25}
\newpage
\input{checklist}
\newpage
\input{appendix}

\end{document}

%% file: intro.tex
\section{Introduction}

Evictions can have a profound impact on tenants, causing instability in the rental market and exacerbating the already significant crisis of affordable housing and homelessness in many large urban areas.
While the response to the COVID-19 pandemic in the United States was to impose moratoria on evictions at federal, state, and local levels, these have now been lifted.
Moreover, most of the \$46 billion allocated in housing assistance for low-income households through the Emergency Rental Assistance (ERA) program has now been spent.
As a result, eviction rates in the US are rising, with an average of 3.6 million eviction cases filed annually \cite{gromisEstimatingEvictionPrevalence2022a,marcal2023}.
Eviction concerns are especially acute due to the inequitable impact on marginalized communities. Female, Black, and families with children disproportionately experience eviction \cite{collinsonEvictionPovertyAmerican2022,graetz2023comprehensive}. Exposure to unstable and substandard housing can be particularly hard on children, leading to developmental effects that can follow them for the rest of their lives \cite{desmondForcedRelocationResidential2015}.

While one way to mitigate the risks and consequences of evictions is through policy, a complementary approach canvasses households at risk of eviction to provide resources to help tenants avoid it.
For example, providing information about the availability of effective legal representation can be instrumental, as far fewer tenants than landlords have legal representation in eviction proceedings~\cite{desmond2016evicted}.
Legal representation for renters has a significant impact, reducing the likelihood of eviction warrants and possessory judgments while also imposing smaller monetary judgments~\cite{cassidyEffectsLegalRepresentation2023,greinerLimitsUnbundledLegal2013a,seronImpactLegalCounsel2001}.
In addition, subsidies that help low-income tenants with utility payments~\cite{utilityAssistance}, along with low-income housing programs such as Section 8 vouchers~\cite{section8}, can improve their ability to pay rent and avoid eviction.
Proactively providing information about these programs can thus also significantly mitigate eviction risk.

Canvassing tenants at risk of eviction, however, is labor intensive. Canvassers working individually or in teams struggle to reach the vast number of low-income housing units behind on rent. It is, consequently, crucial to make efficient use of such limited resources to provide the maximum benefit possible. However, another important challenge is that we do not know, a priori, the risk of eviction for any given household.
We can use past data to learn a predictive model for eviction risk, as demonstrated by \citet{mashiat2024beyond}.
However, such predictions can rapidly become stale, and data we can use to train is not equally available everywhere.
Thus, we need to effectively use a limited canvassing budget to identify at-risk households while improving the quality of predictions we make in identifying such households.
This ultimately necessitates effectively trading off exploration, which allows us to improve predictions of households likely to be evicted and exploitation aimed at reaching the most at-risk households.

We introduce a novel \emph{active geospatial search (AGS)} framework to model this problem.
In AGS, an agent (e.g., canvasser) has a limited budget $\mathcal{C}$ that can be used to query a series of locations (rental units, buildings) embedded in a geographical area.
Each query returns a signal whether or not the location has a target property (an impending eviction, high eviction risk, etc), but incurs a cost which may depend on the previous query (for example, representing travel time between two locations).
The goal in AGS is to find a search policy to maximize the total number of locations with the target property (henceforth, targets) identified within the limited search budget.
Additionally, some eviction data is available---for example, from past court filings, or evictions in a specific geographical region---but not directly applicable to the search problem at hand.
For example, we may have data for past evictions but need to identify \emph{impending} evictions (effectively, evictions that have yet to occur).

AGS builds conceptually on two closely related models: conventional active search and visual active search.
In conventional active search~\cite{garnett2015introducing,jiang2017efficient,jiangCostEffectiveActive2019}, one sequentially queries labels for a given dataset of inputs, but no prior labeled data or relationship topology that relates inputs to one another in a semantically meaningful way is provided.
Thus, typical approaches use myopic and non-myopic heuristics for balancing exploration (to learn a model that predicts labels given inputs) and exploitation coupled with relatively simple predictive models (such as $k$-nearest neighbors).
Visual active search (VAS)~\cite{sarkar2022visual,sarkar2023partially} was recently developed to address active search in which queries are associated with small regions within a large-scale overhead image, with labels corresponding to the existence of a target object within the region.
However, the VAS model is intimately tied to a visual representation of the search area, and therefore, solutions to this problem cannot be directly applied in the AGS setting.


We propose a hierarchical reinforcement learning (RL) approach for scalable AGS.
Our first step is a key building block: composing prediction and search modules, with the latter trained using RL loss, while the parameters of the prediction module are updated using the supervised loss using labels obtained from queries.
The technical challenge is that this approach scales poorly as we consider large geographical areas with thousands of parcels (as our experiments demonstrate).
To address this, we propose a hierarchical policy and learning framework, \emph{hierarchical AGS (HAGS)}.
In HAGS, the area is divided into regions.
We then learn a shared region-level prediction module, a shared region-level search policy that determines the next parcel to query within a relatively small geographical region (as in the first approach), and the high-level policy, which is trained to select which region to query.

We evaluate the proposed approach using eviction data for a large urban area.
First, we show that in settings with uniform query costs and those in which query costs depend on inter-location distance, HAGS outperforms all baselines, including the reinforcement learning approach for small-area AGS, often by a large margin.
Second, we 
show that structural (tabular) features are slightly more useful in isolation than overhead images of parcels, and the combination provides a tangible improvement, demonstrating the value of multimodal information in this context.






In summary, we make the following contributions:
\begin{itemize} 
    \item We propose a novel model of geospatial exploration, \emph{active geospatial search (AGS)}, motivated by the problem of mitigating eviction risk.
    \item We develop an end-to-end deep reinforcement learning pipeline to solve AGS in small-area search problems.
    \item We develop a hierarchical framework to tackle AGS, \emph{HAGS}, in large-area search problems.
    \item We demonstrate the efficacy of HAGS using eviction data from a large urban area, showing that it outperforms all baselines, including conventional active search and a naive application of RL for small-area AGS.
\end{itemize}


%% file: rel_work.tex
\section{Related Work}

Our work is part of a larger body of literature focusing on geospatial applications of optimization and artificial intelligence in nonprofit and humanitarian domains. This includes developing solutions for collaborative recycling \cite{hemmelmayrPeriodicLocationRouting2017}, the sequential redistribution of food donations \cite{balcikMultivehicleSequentialResource2014}, the routing of disaster relief \cite{delatorreDisasterReliefRouting2012}, predicting micronutrient deficiency \cite{bondi-kellyPredictingMicronutrientDeficiency2023}, and anti-poaching measures \cite{fangDeployingPAWSField2016,fang2015security,bondi2018airsim,bondiBIRDSAIDatasetDetection2020,bondiSPOTPoachersAction2018,xuStayAheadPoachers2020}.
However, none of these modeling and solution approaches can be directly applied to the AGS framework in mitigating eviction risk through canvassing and information distribution.


\paragraph{Active Search}

AGS builds on conventional \emph{active search}, first proposed by \citeauthor{garnettBayesianOptimalActive2012}. Previous work in active search has focused on developing nonmyopic algorithms~\cite{jiangEfficientNonmyopicActive2017}, minimizing the cost to find a given number of examples of the target class~\cite{jiangCostEffectiveActive2019}, and extending nonmyopic solutions to multifidelity~\cite{nguyenNonmyopicMultifidelityAcitve2021} and multiclass~\cite{nguyenNonmyopicMulticlassActive2023} settings. 
Recently, \emph{visual active search (VAS)} has been proposed as a variation of active search in which the search region is a satellite image~\cite{sarkar2022visual,sarkar2023partially}.
However, VAS is focused exclusively on visual data and is therefore not directly applicable to AGS.

\paragraph{Geospatial Applications of Visual Data}
 
Geospatial information linked with images has proved useful for the dynamic modeling of traffic \cite{workmanDynamicTrafficModeling2020} and the enhancement of near/remote sensing \cite{workmanRevisitingRemoteSensing2022}. The use of imagery as a source of property information is motivated by the work of \citeauthor{leePredictingGeoinformativeAttributes2015}~\shortcite{leePredictingGeoinformativeAttributes2015} who use images from Flickr to predict geo-informative attributes of the location being photographed, \citeauthor{gebruUsingDeepLearning2017}~\shortcite{gebruUsingDeepLearning2017} who estimate socioeconomic characteristics of neighborhoods based on Google Street View images, and \citeauthor{archboldFineGrainedPropertyValue2023}~\shortcite{archboldFineGrainedPropertyValue2023} who develop fine-level estimates of property value at the pixel level from overhead images.

\paragraph{Eviction and Tenant Harassment Prediction}

Work by \citeauthor{yeUsingMachineLearning2019}~\shortcite{yeUsingMachineLearning2019} and \citet{mashiat2024beyond} in the housing domain has shown promise in utilizing machine learning to predict tenant harassment, but lacks the utilization of high-dimensional visual information as well as a sequential decision-making policy. On the other hand, \citeauthor{tabarWARNERWeaklySupervisedNeural2022}~\shortcite{tabarWARNERWeaklySupervisedNeural2022} have used satellite imagery data to predict whether a given census tract is an eviction hot spot for the county in which it sits, but their model yields a high-level picture of eviction risk, with census tracts covering 4,000 people on average. Other efforts to harness data science methods to predict and understand evictions include forecasting the number of tenants at risk of formal eviction in the next month in a census tract \cite{tabarForecastingNumberTenants2022}, and understanding the predictors of eviction and future eviction hot spots in San Francisco \cite{tanUsingMachineLearning2020}.


%% file: model.tex
\section{Active Geospatial Search for Eviction Prevention}

In this work, we consider the problem of discovering properties with tenants at risk of an upcoming eviction filing, with the goal of reducing this risk, for example, by providing information about financial and legal resources. 
We model this as a \emph{active geospatial search (AGS)} problem.
At the high level, AGS involves sequential exploration and discovery, with the ultimate goal of identifying as many locations with a pre-specified target property as possible given limits on time and resources.
Formally, a geospatial search task consists of a set of $K$ parcels (e.g., rental buildings) embedded as points in a geographic region. 
Each parcel $i$ is associated with a feature vector $x_i$ as well as a geospatial location $l_i \in \mathbb{R}^2$ (while our focus is on 2D geographic coordinates, generalization to 3D coordinates is immediate).
Attributes in $x_i$ can include visual data (such as satellite imagery) as well as tabular data (such as the number of units in the building, year built, and so on).
Let $x = (x_1, \ldots x_i, \ldots, x_K)$ aggregate all of this parcel-level attribute information.
Each parcel $i$ is also associated with a binary label $y_i \in \{0,1\}$, where $y_i = 1$ iff parcel $i$ has the property of interest (e.g., a likely eviction filing in the near future, for example, over the next three months).
Let $y = (y_1, \ldots, y_K)$ denote the vector of labels over all parcels.


A central feature of AGS is that at the beginning of the search, we have label information for a subset of parcels obtained, for example, using a recent history of evictions.
For the rest, our task amounts to both learning (exploration) and discovery (exploitation).
Specifically, we generate a sequence of location queries $\{q_t\}$, where each $q_t$ queries a label $y_{i}$ at location $i=q_t$.
Let $c(i, j)$ as the cost associated with querying parcel $j$ when initiating the query process from parcel $i$. 
To account for the initial query, we introduce a dummy starting parcel $d$, where $c(d, k)$ is the initial query cost.
Let $\mathcal{C}$ be the query budget constraint.
The objective of AGS is to identify as many target parcels as we can within the total budget constraint, which we represent as the following optimization problem:
\begin{equation}
\label{E:objective}
    \max_{\{q_t\}} \ \sum_{t} y_{q_t \quad} \mathrm{s.t.:} \quad  \sum_{t\ge 0} c(q_{t-1},q_t) \le \mathcal{C}
\end{equation}
where $c(q_{-1},q_0) = c(d,q_0)$ is the cost of the first query.

%% file: method.tex
\section{Proposed Approach}




\begin{figure}
  \centering
  \includegraphics[width=0.85\linewidth]{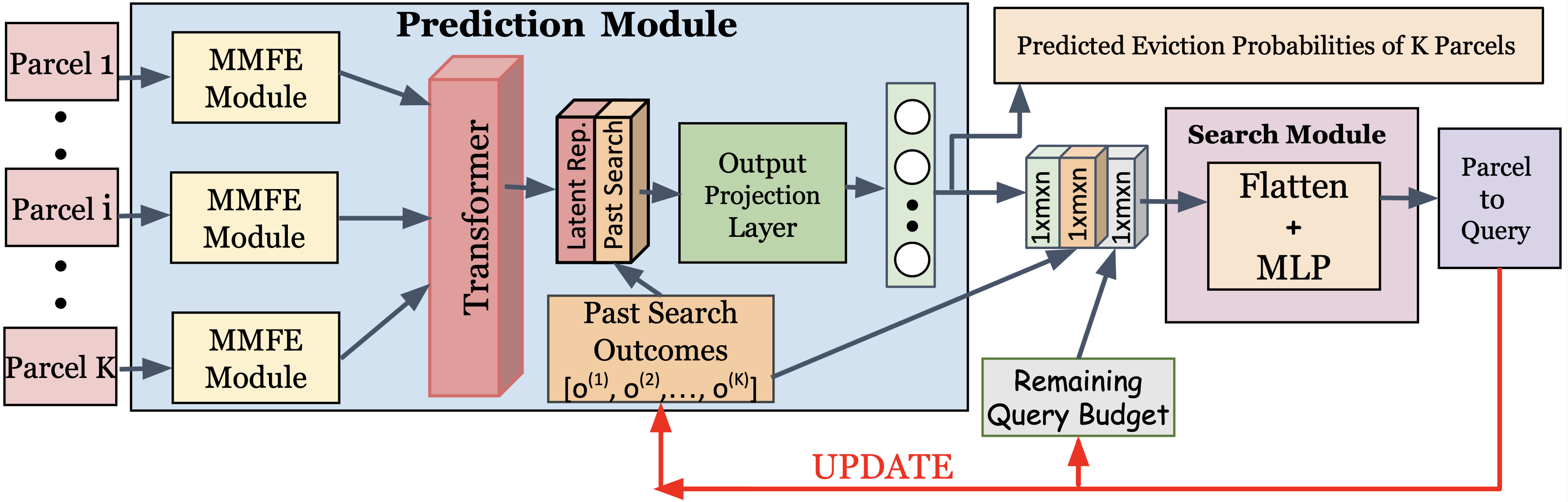}
  \caption{Policy network architecture.}
  \label{fig:policy_net_small}
\end{figure}
\vspace{-1mm}
We begin by considering AGS in a small area; this will provide key building blocks for addressing large-area AGS that we deal with below.
Specifically, we propose an approach for \emph{learning} a search policy from past query results for a subset of locations, which is then deployed to solve Problem~\eqref{E:objective}, balancing exploration (using queries to improve our ability to predict likely target locations) and exploitation (identifying locations that are subject to eviction proceedings, either ongoing or impending).
Our first step to this end is to model AGS as a budget-constrained Markov decision process (MDP), akin to \citeauthor{sarkar2023partially}~\shortcite{sarkar2023partially}.
In this MDP, the \textbf{input state} at time $t$ includes: 1) aggregated feature vectors of the $K$ parcels, $x^t$, which are crucial in providing a broad perspective on the current search state, 2) the outcomes of past search queries $o^t$, and 3) the remaining budget $B^t \leq \mathcal{C}$. 
We represent outcomes of search query history $o$ as follows. Each element of $o$ corresponds to a parcel index $i$, so that $o^t = (o_{t1}, \ldots, o_{tK})$, where $o_{ti} = 0$ if $i$ has not been previously queried, and $o_{ti} = 2 (y_i) - 1$, if parcel $i$ has been previously queried.

In this MDP, the \textbf{actions} are choices over which parcels to query next.
In particular, we denote the set of parcels by $A = \{1, . . . , K\}$. Since, in our model, there is never any value to querying a parcel more than once, we restrict actions available at each step to only parcels that have not yet been queried. 
We assign an \textbf{immediate reward} for query a parcel $i$ as $R(x,i) = y_i$.
Finally, \textbf{state transitions} involve updating the remaining budget by subtracting the current query cost
and incorporating the result of the most recent search query into the outcomes of past search queries.

We begin by proposing a reinforcement learning (RL) approach for learning a search policy when the set of locations available (that is, $K$) is small, limiting the number of actions our search needs to consider.
However, this approach fails to scale to large geographical regions.
Therefore, we subsequently tackle the scalability challenge by proposing an approach for learning \emph{hierarchical} search policies.

\subsection{Small-Area Search}

Suppose that we consider a relatively small geographical area so that the total number of parcels $K$, and therefore, the number of actions $|A|$ that we need to consider, is relatively small.
We propose a RL approach for solving this problem.
Specifically, we use the REINFORCE algorithm to directly learn a search policy $\psi_{\theta}(x, o, B)$,  where $\theta$ are the parameters of the policy that we learn~\cite{williams1992simple}.

In order to utilize the information we acquire during search, following \citeauthor{sarkar2023partially}~\shortcite{sarkar2023partially}, we propose a search policy comprised of two key components: 1) the prediction module represented by $f_{\phi}(x, o)$ and 2) the search module denoted as $g_{\zeta}(p, o, B)$, where $\phi$ and $\zeta$ represent trainable parameters and $p = f_{\phi}(x, o)$ is the vector of predicted eviction probabilities with $p_i$ the predicted probability of at least one eviction in parcel index $i$. 
Conceptually, $f_{\phi}$ generates predictions by exclusively considering the task features $x$ and previous search outcomes $o$, whereas $g_{\zeta}$ depends solely on information pertinent to the search process itself, including the predicted eviction probabilities $p$, previous search outcomes $o$, and the remaining budget $B$. The resulting search policy is a combination of these modules, expressed as $\psi(x, o, B) = g_{\zeta}(f_{\phi}(x, o), o, B)$ (Fig.~\ref{fig:policy_net_small}).


Throughout the episode, we keep the search module $g_{\zeta}$ fixed, using it to generate a sequence of queries, which inherently incorporates an element of exploration due to the stochastic nature of the policy. As we observe labels $y_{j}$ for each queried parcel $j$ during the episode, we update the prediction function $f_{\phi}$ using binary cross-entropy loss ($\mathcal{L}_{BCE}$). Once the episode concludes (when we have exhausted the search budget $C$) we update both the search policy parameters $\zeta$ and the initial prediction function parameters $\phi$. This update involves a combination of RL and supervised loss. In the case of the search module, we calculate the cumulative sum of rewards $R = \sum_{i} y_i$ for the parcels $i$ queried during the episode, and employ the RL loss $\mathcal{L}_{RL}$ based on the REINFORCE algorithm. For the prediction module, we utilize the collected labels $y_{i}$ from the episode and apply $\mathcal{L}_{BCE}$ loss. The proposed approach explicitly balances the RL and supervised loss through the loss function:
\begin{equation}
  \mathcal{L}_{AGS} = (\mathcal{L}_{RL} + \lambda \mathcal{L}_{BCE}).
\end{equation}
This ensures that the policy is trained \emph{to adapt to the evolving prediction dynamics during the episode}. Here $\lambda$ is a hyperparameter. A detailed presentation of the complete method is provided in Algorithm 2 in Supplement. During the inference phase, we fix the parameters of the search module $\zeta$, and udpate the parameters of the prediction module $\phi$ after each query outcome is observed using the $\mathcal{L}_{BCE}$ loss. 

\begin{figure*}
  \centering
  \includegraphics[width=0.85\linewidth,height=4.3cm]{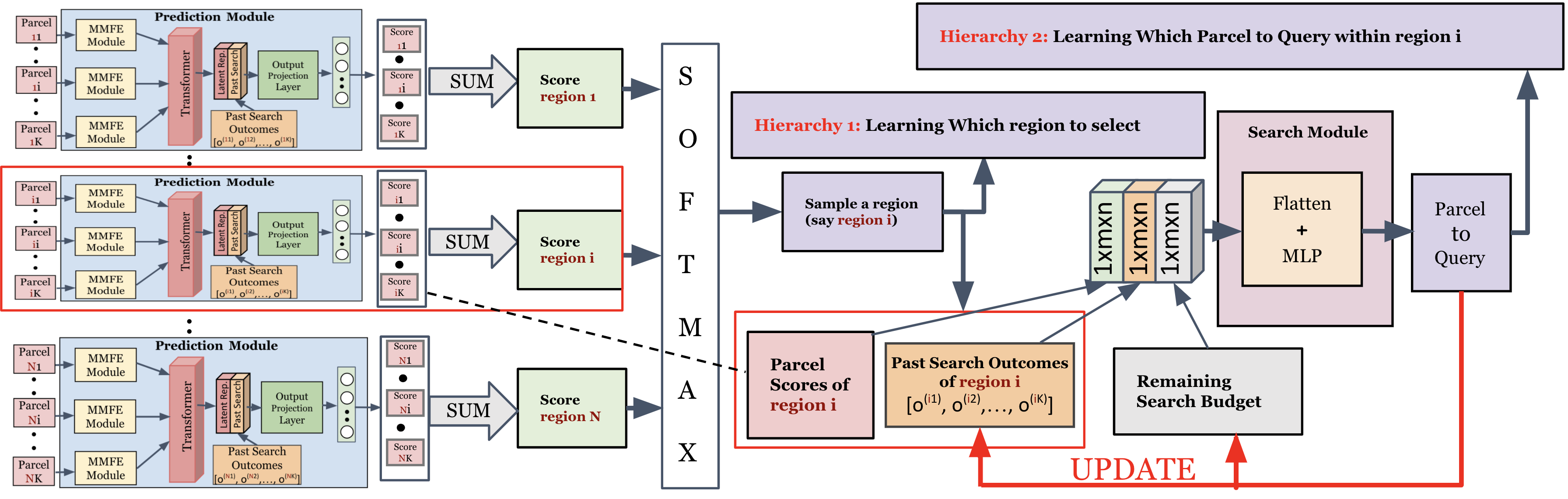}
  \caption{HAGS policy network architecture.}
  \label{fig:policy_net}
\end{figure*}

\subsection{Large-Area Search}

The key assumption in the approach above is that the number of candidate parcels is relatively small.
In practice, that is unrealistic, since even a reasonable target geographical area may contain tens of thousands of parcels.
Since the architecture described above requires a policy output per action (parcel), it cannot scale to such problems (see Section~\ref{S:exp}).

To address this issue, we propose a hierarchical search framework, 
\textit{Hierarchical AGS (HAGS)}. 
The key insight behind HAGS is that we can leverage shared structure---in particular, symmetry and geospatial locality---of the geospatial domain to introduce inductive bias that significantly reduces learning and decision complexity.

Specifically, let the geospatial area of interest be comprised of $N$ regions, where each region is, in turn, comprised of (at most) $K$ parcels.
This induces a hierarchical decomposition of the area first into regions (first level), and then (within each region) into parcels (second level).
In HAGS, the first level of decision making will therefore correspond to choosing a region, while the second will entail choosing a parcel within the selected region.
Consequently, a level-1 (higher-level) policy will choose among the $N$ regions, whereas a level-2 policy for each region $r$ will, in turn, choose among the $K$ parcels.
For a region $r$, let $x_r = (x_{r1},\ldots,x_{rK})$ denote the collection of attributes for each parcel in $r$, and let $x = (x_1,\ldots,x_N)$ aggregate all of these into a single global feature vector over all regions.
Similarly, $o_r$ is a vector of observed query responses over the parcels in region $r$, with $o$ combining them into a single global vector.

In HAGS, as in our approach for small-area search above, we decompose the search problem into two pieces: 1) a prediction module $f_\phi(x_r, o_r)$ which outputs parcel-level predictions given region-level inputs $x_r$ and $o_r$, and 2) a hierarchy of search policies, as visualized in Figure~\ref{fig:policy_net}.
The main idea in our HAGS architecture is to leverage geospatial structure by a) learning a single prediction module $f_\phi(x_r, o_r)$ with parameters $\phi$ shared across both the first and second levels of decision making, and b) learning a single level-2 policy $g^{h2}_\theta(x_r,o_r,B)$ shared by all regions.
This introduces an inductive bias, taking advantage of geospatial structure to significantly reduce the number of parameters we need to learn.

Specifically, let $p=f_\phi(x,o)$ denote predictions for \emph{all} regions, aggregated from individual region-level predictions, with $p_{ri}$ the predicted probability that parcel $i$ in region $r$ has the target property (e.g., an impending eviction proceedings).
Let $\bar{p}_r = \sum_{i} p_{ri}$ be the predicted expected number of parcels with the target property in region $r$ ( where $i$ ranges over parcels in region $r$).
Let $g^{h1}_\zeta(\bar{p},o,B)$ denote the level-1 policy that outputs a distribution over regions to query next, given inputs $p$, $o$ (aggregated over all regions) and remaining budget $B$.
Similarly, let $g^{h2}_\theta(p_r,o_r,B)$ be a shared level-2 policy (shared across all regions) that given region-specific inputs $p_r$, $o_r$, along with $B$ outputs a distribution over parcels in the associated region.
At training time, actions are sampled from these distributions, whereas at search time we choose the action with the highest probability.


We jointly train the parameters of the prediction module, as well as both the level-1 and level-2 policies, using the REINFORCE policy gradient framework, with the loss function 
\begin{equation}
  \mathcal{L}_{HAGS} = (\mathcal{L}^{h1}_{RL} + \mathcal{L}^{h2}_{RL} + \lambda \mathcal{L}_{BCE}),
\end{equation}
where $\mathcal{L}^{h1}_{RL}$ and $\mathcal{L}^{h2}_{RL}$ are the standard REINFORCE loss functions for level-1 and level-2 policies, respectively, and $\mathcal{L}_{BCE}$ the supervised binary cross-entropy loss used to train the prediction module.
The RL rewards for the level-2 policies are just as in AGS, that is, 1 if the queried parcel has the target property and 0 otherwise.
For level-1 policies, the reward associated with a chosen parcel is 1 if the query according to the level-2 policy within this region yields the target property, and 0 otherwise.
Note that $\mathcal{L}_{BCE}$ plays an identical role as it did in small-area search AGS framework, that is, we dynamically modify the parameters of the prediction module ($\phi$) following the observation of each query outcome both during training and inference. This adjustment is achieved through the utilization of the $\mathcal{L}_{BCE}$ loss, which computes the binary cross-entropy loss between the predicted label and the observed label $y$ for the queried parcel.
During gradient descent steps, we backpropagate $\mathcal{L}^{h1}_{RL}$ and $\mathcal{L}_{BCE}$ through the prediction module and backpropagate $\mathcal{L}^{h2}_{RL}$ through the level-2 policy, updating parameters of both $f$ and the associated search policy.
A detailed formal presentation of the method is provided in Algorithm~\ref{A:MASHA-Hierarchical}.

\begin{algorithm}[!h]
\scriptsize
\caption{The proposed \textsc{HAGS} algorithm for training.}
\begin{algorithmic}[1]
\scriptsize
\Require  A search task instance $(x = [x_1,\ldots, x_N]; y = [y_{1}, \ldots y_{r}, \ldots,  y_{N}])$, where $x_{r} = [x_{r1}, \ldots, \ldots, x_{rK}]$ and $y_{r} = [y_{r1}, \ldots,  y_{rK}]$ ; budget constraint $\mathcal{C}$; Hierarchy 1 policy: $g^{h1}_\zeta(p = f_{\phi}(x,o), B)$ with parameters $\zeta$ and prediction module $f$ parameterized by $\phi$; Hierarchy 2 policy: $g^{h2}_\theta(p_r, o_r, B)$ with parameters $\theta$; $o = [o_1,\ldots, o_N]$ with $o_r = [o_{r1},\ldots, o_{rK}]$;  \\
\STATE \textbf{Initialize} $o_{r} = [ 0...0]$ for each $r \in \{1, \ldots, N\}$;  $B^{t} = \mathcal{C}$; step $t = 0$
\While {$B^{t} > 0 $}
    \State $p = f_{\phi}(x,o)$; here $p = [p_1,\ldots, p_N]$ with $p_{r} = [p_{r1}, \ldots,  p_{rK}]$
    \State \emph{j} $\xleftarrow{} \mathit{Sample}_{j \in \{\mathit{1, \ldots r, \ldots, N}\}} \mathit{Softmax}[\bar{p}]$; here $\bar{p} = [\bar{p}_1,\ldots, \bar{p}_N]$ and $\bar{p}_r = \sum_{i} p_{ri}$ with $i \in \{1, \ldots, K\}.$.
    \State Explore region with index $j$ at time t.
    \State $\tilde{s}$ = $g^{h2}_\theta(p_j, o_j, B^t)$
    \State \emph{r} $\xleftarrow{} \mathit{Sample}_{r \in \{\mathit{1, \ldots i, \ldots, K}\}} [\tilde{s}]$
    \State Query parcel with index $r$ within region $j$ and observe true label $y_{jr}$.
    \State Update $\phi^{t}$ to $\phi^{t+1}$ using $\mathcal{L}_{\mathit{BCE}}$ loss between $p$ and pseudo label $\hat{y}^{t}$, each component of $\hat{y}^{t}$ is defined as \\  
    $\hat{y}^{t}_{jr}\xleftarrow{}\left\{
     \begin{array}
     {r@{\quad}l}
          \text{$y_{jr}$} & \text{if $r$'th parcel within region $j$ has been queried } \\
          \text{$p_{jr}$}  & \text{if  $y_{jr}$ is Unobserved.} 
     \end{array}
     \right.$
    \State Set $R^t = y_{jr}$, Update $o^t$ to $o^{t+1}$  with $o_{jr} = 2y_{jr} -1$, update $B^{t}$ to $B^{t+1}$ with $B^{t+1} = B^{t} - c(k,j)$ where $k$ is the parcel queried at $t-1$.
    \State Collect transition tuple $\tau$ at step t, i.e., $\tau^{t}$ = $\bigl(\text{state} = (x, o^t, B^{t})$, action of hierarchy 1 policy = $j$, action of hierarchy 2 policy = $r$, reward of both policies = $R^{t}$, next state of hierarchy 1 policy = $(x, o^{t+1}, B^{t+1})$, next state of hierarchy 2 policy = $(p_{z}, o_{z}^{t+1}, ,B^{t+1})\bigr)$ assuming level 1 policy selects region $z$ at (t+1). 
    \State \emph{t} $\xleftarrow{} t+1$
\EndWhile \\
\STATE Update hierarchy 1 policy parameters using ($\mathcal{L}_{\mathit{RL}}^{\mathit{h1}}$) based on the collected transition tuples ($\tau^{t}$) throughout the episode and also update the prediction module parameters $\phi$ using ($\mathcal{L}_{\mathit{BCE}}$) based on the collected labels ($y_{jr}$) over the episode.\\
\STATE Update hierarchy 2 policy parameters $\theta$ using ($\mathcal{L}_{\mathit{RL}}^{\mathit{h2}}$) based  on the collected transition tuples ($\tau^{t}$) throughout the episode.\\
\STATE \textbf{Return} Updated hierarchy 1 and 2 policy parameters.
\end{algorithmic}
\label{A:MASHA-Hierarchical}
\end{algorithm}

%% file: results.tex
\section{Experiments}
\label{S:exp}

\begin{table*}
\vspace{-2mm}
    \centering
    \footnotesize
    \caption{\small{Average number of targets (ANT) found as a function of search budget for uniform query cost; average eviction rate is 5\%.}
    }
    \begin{tabular}{p{3cm} *{12}{p{0.75cm}}}
        \toprule
        Search Budget & $15$ & $20$ & $25$ & $50$ & $75$ & $100$ & $200$ & $300$ & $400$  \\
        \midrule
        Random & 1.00 & 1.75 & 2.00 & 2.25 & 2.50 & 2.75 & 9.75 & 12.50 & 21.75  \\
        \small{Conventional AS} & 2.25 & 3.25 & 4.50 & 5.75 & 6.50 & 7.50 & 15.75 & 22.00 & 32.50 \\
        \small{Greedy by Unit Count} & 5.25 & 5.50 & 7.00 & 12.50 & 20.00 & 25.75 & 36.25 & 46.00 & 55.25  \\
        Greedy & 7.50 & 9.25 & 10.50 & 17.75 & 21.00 & 29.50 & 44.50 & 59.00 & 65.75  \\ 
        Greedy Adaptive & 8.75 & 10.75 & 14.00 & 22.25 & 27.50 & 34.75 & 54.50 & 65.50 & 72.25  \\ 
        AGS & 8.00 & 10.25 & 11.50 & 18.75 & 22.00 & 31.50 & 46.25 & 61.25 & 68.25  \\ 
        \textbf{HAGS} & \textbf{10.25} & \textbf{12.75} & \textbf{14.50} & \textbf{25.75} & \textbf{29.50} &\textbf{38.50} & \textbf{57.75} & \textbf{70.50} & \textbf{77.75}  \\
        \bottomrule
    \end{tabular}
    \label{tab:large_Scale_5}
\end{table*}

\begin{table*}[!h]
\vspace{-2mm}
    \centering
    \footnotesize
    \caption{\small{Average number of targets (ANT) found as a function of search budget for uniform query cost; average eviction rate is 10\%.}
    }
    \begin{tabular}{p{3cm} *{12}{p{0.75cm}}}
        \toprule
        Search Budget & $15$ & $20$ & $25$ & $50$ & $75$ & $100$ & $200$ & $300$ & $400$   \\
        \midrule
        Random & 1.25 & 2.25 & 2.75 & 5.50 & 7.75 & 9.50 & 16.25 & 29.75 & 46.75  \\
        \small{Conventional AS} & 3.50 & 5.50 & 6.75 & 11.25 & 15.50 & 18.00 & 27.50 & 43.50 & 65.50   \\
        \small{Greedy by Unit Count} & 8.00 & 11.25 & 14.25 & 23.25 & 32.25 & 42.25 & 50.00 & 72.00 & 105.50  \\
        Greedy & 9.75 & 12.75 & 17.00 & 28.75 & 38.25 & 51.00 & 74.25 & 103.50 & 119.75  \\ 
        Greedy Adaptive & 10.50 & 14.50 & 18.75 & 34.50 &49.25& 57.50 & 86.75 & 117.75 & 144.50  \\ 
        AGS & 10.50 & 14.00 & 17.25 & 29.25 & 41.50 & 53.25 & 79.25 & 106.75 & 128.25  \\ 
        \textbf{HAGS} & \textbf{13.25} & \textbf{18.00} & \textbf{22.25} & \textbf{38.50} & \textbf{51.75} & \textbf{60.00} & \textbf{91.75} & \textbf{124.50} & \textbf{151.75}  \\
        \bottomrule
    \end{tabular}
    \label{tab:large_Scale_10}
\end{table*}

We evaluate the efficacy of the AGS framework and the proposed HAGS approach using observed eviction filings in a mid-sized region.
Our \emph{evaluation metric} is the average number of targets (ANT) found within a given budget (averaged over search runs).
We consider two query cost settings: (i) uniform query costs, i.e., $c(i,j)=1$ for all parcels $i,j$, and (ii) distance-based cost, where $c(i,j)$ is proportional to the distance between $i$ and $j$.
Next, we describe in detail the data we use, as well as the baseline methods, before presenting our results.
Our focus here is on large-area search; we defer most results involving small-area search to the Supplement.





\smallskip
\noindent\textbf{Data }
We construct features associated with parcels from two sources: tabular data and overhead (satellite) images.
The tabular features are based on those in \citet{mashiat2024beyond}, and encompass eviction court filings, owner information, property-level attributes, and neighborhood features.
These data are originally derived from a collection of municipal sources across St. Louis City and County, excluding neighborhood features, which are obtained from the American Community Survey (ACS)~\cite{acs2021}.
Court eviction filings are aggregated over the previous year, as well as semiannually, quarterly, and monthly. 
Property information includes the number of housing units and whether it is owner-occupied. 
Properties are linked to owners, and information on linked owners is included, such as the number of properties owned, in- versus out-of-state residence, and whether the owner has worked with a moderate- or high-filing attorney (defined as a filing rate above one and three standard deviations above the mean for attorneys during that period, respectively).
Neighborhood features are at the Census block-group level, and describe areas of 600-3000 households in terms of average rent
as a percentage of household income, the median household income, the ratio of income to the poverty level, racial and ethnic composition, the number of total housing units, the proportion of occupied and vacant units, and the proportion of owner and renter-occupied units.
We restrict our analysis to residential, non-vacant parcels with at least two rental units, yielding a total of 26700 properties across St. Louis City and County.
The time period used for training all AGS and baseline approaches is July 1, 2021, to September 30, 2022.
Testing covers the period between October 1, 2022, and December 31, 2022, where the target is positive if an eviction filing occurred at the property during that period.

Satellite imagery data comes from the National Agriculture Imagery Program (NAIP)~\cite{NationalAgricultureImagery}. Images were captured during June 2022 at a resolution of 60 centimeters. We extracted $214 \times 214$ patches such that these patches fully covered 95\% of the properties.

\begin{figure}
  \centering
  \includegraphics[width=1\linewidth]{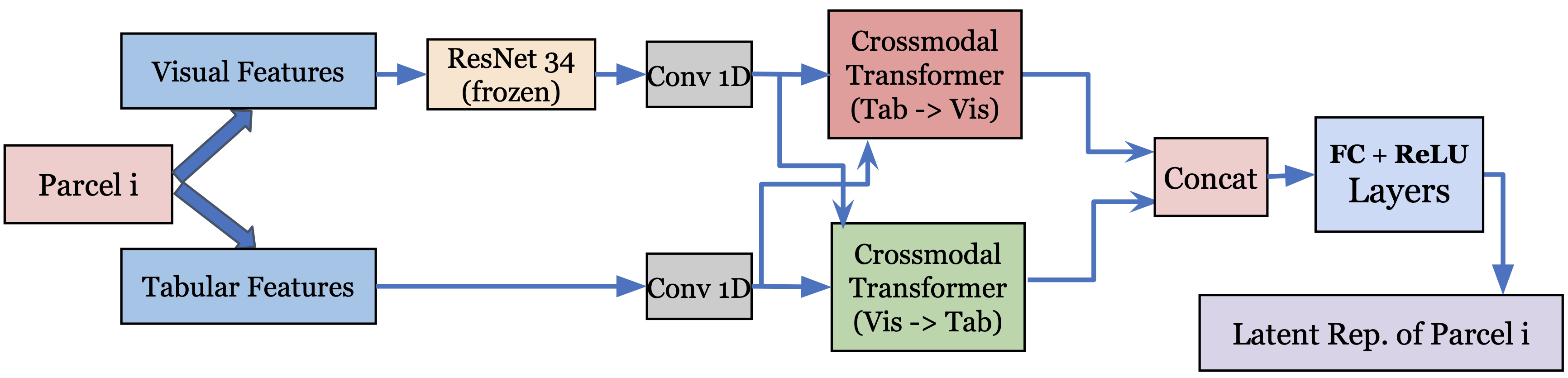}
  \caption{\small{Multi-Modal Feature Extraction module for each Parcel.}}
  \label{fig:fe}
\vspace{-3pt}
\end{figure}

Since we possess visual data for each individual parcel as well as tabular features containing past eviction records for the corresponding parcels, we utilize methods from multi-modal representation learning~\cite{ngiam2011multimodal,baltruvsaitis2018multimodal,tsai2019multimodal}. This enables us to seamlessly amalgamate information from both modalities, culminating in a latent representation for each parcel. We leverage a widely adopted multi-modal transformer architecture as described in~\cite{tsai2019multimodal}. We depict the Multi-Modal Feature Extraction (MMFE) module in Figure~\ref{fig:fe}. 
Note that the parameters of MMFE modules are shared across the parcels. 

\smallskip
\noindent\textbf{Baseline Methods }
We compare the proposed approach to the following baselines:
\begin{enumerate}[noitemsep,topsep=0pt]
\item \emph{Random:} Each parcel is chosen uniformly randomly among those not yet explored.
\item \emph{Greedy:} we train a classifier $f_\mathit{greedy}$ to predict whether a particular property will have at least one eviction filing within the next three months and search the most likely properties until the search budget is exhausted.
\item \emph{Greedy by unit count:} query the parcel with the largest number of units.
\item \emph{Greedy adaptive:} similar to \emph{Greedy} except the prediction model $f$ is updated at each step based on query outcomes.
\item \emph{Conventional active search}, an active search method by \citeauthor{jiang2017efficient}~\shortcite{jiang2017efficient}, using a low-dimensional feature representation for each parcel from the same MMFE feature extraction network as in our approach.
\end{enumerate}
In addition, we use the simple small-area AGS approach as a baseline in the large-area search setting (involving 16000 parcels; see below for further details).

\begin{table*}
\vspace{-2mm}
    \centering
    \footnotesize
    \caption{
    \small{Average number of targets (ANT) found as a function of search budget for distance-based query cost; average eviction rate is 5\%.}
    }
    \begin{tabular}{p{3cm} *{12}{p{0.75cm}}}
        \toprule
        Search Budget & $300$ & $600$ & $1200$ & $2400$ & $4800$ & $10000$ & $20000$ & $40000$ & $80000$ \\
        \midrule
        Random & 0.00 & 0.25 & 0.25 & 0.50 & 0.50 & 0.75 & 0.75 & 1.00 & 2.25 \\
        \small{Conventional AS} & 0.00 & 0.00 & 0.25 & 0.75 & 1.00 & 1.00 & 1.25 & 1.50 & 2.00 \\
        \small{Greedy by Unit Count} & 0.00 & 0.00  & 0.00 & 1.00 & 1.25 & 1.25 &  1.75 & 2.00 & 2.75 \\
        Greedy & 0.00 & 0.00 & 0.00 & 0.00 & 0.00 & 1.00 & 1.00 & 1.75 & 2.75 \\ 
        Greedy Adaptive & 0.00  & 0.00 & 0.00 & 0.00 & 0.00 & 1.00 & 1.00 & 1.75 & 2.75 \\ 
        AGS & 1.25  & 1.50 & 1.50 & 1.50 & 1.50 & 1.75 & 1.75 & 2.00 & 2.50 \\ 
        \textbf{HAGS} & \textbf{2.00} & \textbf{2.50} & \textbf{2.50} & \textbf{2.50} & \textbf{2.75} & \textbf{2.75} & \textbf{2.75}  & \textbf{3.50}  & \textbf{4.50}  \\
        \bottomrule
    \end{tabular}
    \label{tab:large_Scale_5_dist}
\end{table*}

\begin{table*}[h!]
\vspace{-2mm}
    \centering
    \footnotesize
    \caption{
    \small{Average number of targets (ANT) found as a function of search budget for distance-based query cost; average eviction rate is 10\%.}
    }
    \begin{tabular}{p{3cm} *{12}{p{0.75cm}}}
        \toprule
        Search Budget & $300$ & $600$ & $1200$ & $2400$ & $4800$ & $10000$ & $20000$ & $40000$ & $80000$ \\
        \midrule
        Random & 0.00 & 1.00 & 1.00 & 1.25 & 1.25 & 1.25 & 1.50 & 1.75 & 2.25 \\
        \small{Conventional AS} & 0.00 & 0.00 & 0.50 & 1.00 & 1.75 & 1.75 & 2.00 & 2.00 & 2.75 \\
        \small{Greedy by Unit Count} & 1.00 & 1.00 & 1.00 & 1.00 & 1.50 & 1.75 & 2.25 & 2.75 & 3.75 \\
        Greedy & 1.25 & 1.25 & 1.50 & 1.50 & 1.50 & 2.00 & 2.50 & 3.00 & 3.75\\ 
        Greedy Adaptive & 1.25 & 1.25 & 1.50 & 1.50 & 1.50 & 2.00 & 2.50 & 3.00 & 3.75 \\ 
        AGS & 1.50 & 1.75 & 1.75 & 1.75 & 1.75 & 2.00 & 3.00  & 3.50 & 4.25 \\ 
        \textbf{HAGS} & \textbf{2.25} & \textbf{2.50} & \textbf{2.75} & \textbf{2.75} & \textbf{3.00} & \textbf{3.25} & \textbf{4.00}  & \textbf{5.75}  & \textbf{6.25}  \\
        \bottomrule
    \end{tabular}
    \label{tab:large_Scale_10_dist}
\end{table*}
\vspace{-1mm}
\begin{table*}[!h]
\vspace{-2mm}
    \footnotesize
    \centering
    \caption{\small{Ablation Study Average Number of Targets (ANT) Found by Search Task Parameters and Solution Method}}
    \begin{tabular}{p{3.5cm} *{9}{p{1cm}}}
        \toprule
        \multicolumn{1}{c}{} & \multicolumn{3}{c}{Average Positive Rate of 2.5\%} & \multicolumn{3}{c}{Average Positive Rate of 5\%} & \multicolumn{3}{c}{Average Positive Rate of 10\%} \\
        \midrule
        Search Budget & $15$ & $20$ & $25$ & $15$ & $20$ & $25$ & $15$ & $20$ & $25$\\
        \midrule
        Random  & 0.328 & 0.472 & 0.652 & 0.656 & 0.856 & 1.276 & 1.428 & 1.940 & 2.428 \\
        AGS-VIS & 0.932 & 1.092 & 1.268 & 1.840 & 2.216 & 2.532 & 3.496 & 4.100 & 5.044 \\
        AGS-TAB & 1.008 & 1.184 & 1.340 & 1.972 & 2.336 & 2.640  & 4.216 & 5.004 & 5.501 \\ 
        \textbf{AGS} & \textbf{1.052} & \textbf{1.288} & \textbf{1.352} & \textbf{2.240} & \textbf{2.596} & \textbf{2.828} & \textbf{4.372} & \textbf{5.204} & \textbf{5.632} \\
        \bottomrule
    \end{tabular}
    \label{tab:ablation_study}
\vspace{-3pt}
\end{table*}
\vspace{-2mm}

\subsection{Results}
\vspace{-3pt}
Starting with the original 26700 parcels, we randomly select 16000 properties for evaluation, averaging the results over four such random selections to compute ANT.
The selected properties are bootstrapped 
to have a particular average number of positive targets (that is, properties with eviction filings during the prediction period) to enable us to study the impact of target sparsity. 
We consider mean positive rates of 5\% and 10\%, with s.t.d. of 0.01\%, and 0.02\%, respectively.
In HAGS, we divide the entire search space into $N=160$ regions, each containing $K=100$ parcels. 







\smallskip
\noindent\textbf{Uniform-Cost Search }
We first consider uniform-cost settings.
In this case, we consider search budgets $\mathcal{C}$ of 15, 20, 25, 50, 75, 100, 200, 300, and 400 queries.
The results are presented in Tables~\ref{tab:large_Scale_5} and \ref{tab:large_Scale_10} for positive rates of 5\% and 10\%, respectively. 
In all cases, HAGS outperforms all baselines, often by a large margin.
In particular, improvement ranges from 3\%-17\% over the most competitive baseline.
Particularly noteworthy is the poor performance of ``flat'' AGS designed for small-area search.
While AGS outperforms most baselines, it nevertheless exhibits poor efficacy compared to HAGS and, indeed, is slightly worse than the simple greedy adaptive search heuristic.
This suggests the inductive bias introduced in the hierarchical architecture of HAGS is crucial to obtaining high efficacy at scale.
We also observe a general pattern of greatest improvement from HAGS in settings with lower budgets (greatest improvement is for $\mathcal{C}=15$), although neither pattern is monotonic. 
Additionally, a lower overall target (eviction) rate has a greater average improvement (average improvement for 5\% target rate is $\sim$7\%, compared to $\sim$5\% for 10\% target rate), although this pattern is not uniform across budgets.
In other words, the general pattern is that the significance of efficient balancing between exploration and exploitation as exhibited by HAGS is most notable when there is scarcity in either budget or availability of targets to discover.

\smallskip
\noindent\textbf{Distance-Based Search Costs }
Next, we consider HAGS compared to baselines in the case when search costs are not uniform, but instead are based on relative distances between parcels.
Specifically, we determine the position of each parcel using GPS coordinates and calculate the query cost as the Manhattan distance in meters between parcel locations.
We vary search budgets
$\mathcal{C} \in \{300, 600, 1200, 2400, 4800, 10000, 20000, 40000, 80000\}$, and again consider 5\% and 10\% positive rate in the area.
The results are presented in Tables~\ref{tab:large_Scale_5_dist} and~\ref{tab:large_Scale_10_dist}.
In this setting, we see an even greater improvement of HAGS over the baselines (including, again, AGS), with improvement over the most competitive baseline ranging from approximately 42\% to 70\%.
Notably, when the average positive rate is low (5\%), the greedy (exploitation-only) baselines have trouble finding \emph{any} targets within the available budget.

\smallskip
\noindent\textbf{Ablation Study of Visual and Structured Data }
Finally, we address an important qualitative question in the particular context of tenant eviction data analytics: to what extent is visual and structured (tabular) data contribute to decision efficacy?
We study this in a small-area search setting, where we randomly select a region containing 100 parcels.
In particular, we train AGS (no need for HAGS here) using visual-only data (\emph{AGS-VIS}), tabular-only data (\emph{AGS-TAB}), and both (standard \emph{AGS}).
We also provide results of random queries for calibration purposes.
The results are provided in Table~\ref{tab:ablation_study}.
First, note that both \emph{AGS-VIS} and \emph{AGS-TAB} outperform random search by a large margin.
Second, the results suggest that while tabular features tend to be more informative than visual features individually in this setting, integrating visual and tabular features through multi-modal representation learning meaningfully (albeit not dramatically) enhances search performance across various search tasks in different settings. 
These results demonstrate the importance of leveraging the multi-modal representation that combines tabular and imagery-based features.
\vspace{-4pt}

%% file: conclusion.tex
\section{Conclusion}

We introduce the novel AGS framework to identify properties with renters who are at risk of imminent eviction. Through extensive experiments, we demonstrate that our approach increases the number of at-risk properties discovered as compared to several strong baselines by at least 5\% and, in some settings, over 50\%. This is achieved through a pretraining phase combined with an exploration phase that allows for test-time adaptation. These methods have the potential to dramatically increase both the effectiveness and timeliness of door-to-door outreach by social service agencies, thereby increasing the number of tenants who are connected with legal aid, landlord mediation, one-time financial assistance, time-limited case management, or moving assistance. This work is a first step in illustrating the potential of AGS strategies to enhance predictive outreach to address a myriad of social phenomena including evictions and beyond.

%% file: checklist.tex
\section{Reproducibility Checklist}
\label{appendix:checklist}
This paper:
\begin{itemize}
    \item Includes a conceptual outline and/or pseudocode description of AI methods introduced (yes)
    \item Clearly delineates statements that are opinions, hypothesis, and speculation from objective facts and results (yes)
    \item Provides well marked pedagogical references for less-familiare readers to gain background necessary to replicate the paper (yes)
\end{itemize}

\vspace{0.6cm}
Does this paper make theoretical contributions? (no)

If yes, please complete the list below.

\begin{itemize}
    \item All assumptions and restrictions are stated clearly and formally. (yes/partial/no)
    \item All novel claims are stated formally (e.g., in theorem statements). (yes/partial/no)
    \item Proofs of all novel claims are included. (yes/partial/no)
    \item Proof sketches or intuitions are given for complex and/or novel results. (yes/partial/no)
    \item Appropriate citations to theoretical tools used are given. (yes/partial/no)
    \item All theoretical claims are demonstrated empirically to hold. (yes/partial/no/NA)
    \item All experimental code used to eliminate or disprove claims is included. (yes/no/NA)
\end{itemize}

\vspace{0.6cm}
Does this paper rely on one or more datasets? (yes)

If yes, please complete the list below.

\begin{itemize}
    \item A motivation is given for why the experiments are conducted on the selected datasets (yes)
    \item All novel datasets introduced in this paper are included in a data appendix. (NA)
    \item All novel datasets introduced in this paper will be made publicly available upon publication of the paper with a license that allows free usage for research purposes. (yes)
    \item All datasets drawn from the existing literature (potentially including authors’ own previously published work) are accompanied by appropriate citations. (yes)
    \item All datasets drawn from the existing literature (potentially including authors’ own previously published work) are publicly available. (yes)
    \item All datasets that are not publicly available are described in detail, with explanation why publicly available alternatives are not scientifically satisficing. (NA)
\end{itemize}

\vspace{0.6cm}
Does this paper include computational experiments? (yes)

If yes, please complete the list below.

\begin{itemize}
    \item Any code required for pre-processing data is included in the appendix. (yes).
    \item All source code required for conducting and analyzing the experiments is included in a code appendix. (yes)
    \item All source code required for conducting and analyzing the experiments will be made publicly available upon publication of the paper with a license that allows free usage for research purposes. (yes)
    \item All source code implementing new methods have comments detailing the implementation, with references to the paper where each step comes from (yes)
    \item If an algorithm depends on randomness, then the method used for setting seeds is described in a way sufficient to allow replication of results. (yes)
    \item This paper specifies the computing infrastructure used for running experiments (hardware and software), including GPU/CPU models; amount of memory; operating system; names and versions of relevant software libraries and frameworks. (yes)
    \item This paper formally describes evaluation metrics used and explains the motivation for choosing these metrics. (yes)
    \item This paper states the number of algorithm runs used to compute each reported result. (yes)
    \item Analysis of experiments goes beyond single-dimensional summaries of performance (e.g., average; median) to include measures of variation, confidence, or other distributional information. (yes)
    \item The significance of any improvement or decrease in performance is judged using appropriate statistical tests (e.g., Wilcoxon signed-rank). (yes)
    \item This paper lists all final (hyper-)parameters used for each model/algorithm in the paper’s experiments. (yes)
    \item This paper states the number and range of values tried per (hyper-) parameter during development of the paper, along with the criterion used for selecting the final parameter setting. (yes)
\end{itemize}

%% file: appendix.tex
\clearpage
\appendix
\section*{Supplementary Material: Active Geospatial Search for Efficient Tenant Eviction Outreach}
\vspace{3pt}
\section{Search Task Design for Small Action Space}
\vspace{3pt}
We construct a series of search tasks, each containing 100 properties. These 100 properties are bootstrapped from the original 26,700 properties to have a particular average number of positive targets (that is, properties with eviction filings during the prediction period). The mean positive rates of 2.5\% (close to the base rate in the data), 5\%, and 10\% with standard deviations of 0.005\%, 0.01\%, and 0.02\%, respectively. Experiments are run for budgets of 15, 20, and 25 queries, and each parameterization is trained and tested on different sets of 250 search tasks.
\begin{table*}[!h]
    \centering
    \footnotesize
    \caption{\small{Average Number of Targets (ANT) Found by Search Task Parameters and Solution Method}}
    \begin{tabular}{p{2.7cm} *{9}{p{1cm}}}
        \toprule
        \multicolumn{1}{c}{} & \multicolumn{3}{c}{Average Positive Rate of 2.5\%} & \multicolumn{3}{c}{Average Positive Rate of 5\%} & \multicolumn{3}{c}{Average Positive Rate of 10\%} \\
        \midrule
        Search Budget & $15$ & $20$ & $25$ & $15$ & $20$ & $25$ & $15$ & $20$ & $25$ \\
        \cmidrule(r){1-1} \cmidrule(r){2-4} \cmidrule(r){5-7} \cmidrule(l){8-10}
        Random & 0.328 & 0.472 & 0.652 & 0.656 & 0.856 & 1.276 & 1.428 & 1.940 & 2.428 \\
        Greedy\>by\>Unit\>Count & 0.820 & 1.024 & 1.216 & 1.616 & 2.072 & 2.444 & 2.924 & 3.784 & 4.580 \\
        Greedy & 0.540 & 0.632 & 0.804 & 1.660 & 2.040 & 2.308 & 3.796 & 4.296 & 4.780 \\ 
        Greedy \large{Adaptive} & 0.556 & 0.716 & 0.828 & 1.788 & 2.124 & 2.400 & 3.872 & 4.452 & 5.020 \\ 
        \textbf{AGS} & \textbf{1.052} & \textbf{1.288} & \textbf{1.352} & \textbf{2.240} & \textbf{2.596} & \textbf{2.828} & \textbf{4.372} & \textbf{5.204} & \textbf{5.632} \\
        \bottomrule
    \end{tabular}
    \label{tab:ant_main}
\end{table*}
\vspace{-4pt}
\paragraph{Search Performance over Small Action Space}
\vspace{4pt}
We evaluate the proposed methods in different settings with varying $\mathcal{C} \in \{ 15, 20, 25\}$ and average positive rates in $\{ 2.5\%, 5\%, 10\%\}$. The results are presented in \ref{tab:ant_main}. We observe a substantial improvement in the performance of AGS compared to all the baseline approaches, ranging from approximately 16\% - 90\% compared to the most competitive approach, \emph{Greedy Adaptive}. We also observe the \emph{Greedy Adaptive} outperforms the \emph{Greedy} baseline method across all experimental settings. This result highlights the importance of inference time adaptive strategies for optimizing search efficiency. We also observe that the performance gap between our proposed method and the baseline approaches is more significant in the case when evictions are scarce and resources are limited. This is especially important as the average positive rate of 2.5\% is close to the original positive rate in the data, which ranges from 1.8\% to 2.4\%. 

\subsection{Details of Training Hyperparameters}

For training the policy network of AGS and HAGS framework, we use a initial learning rate of $1e-4$, batch size of 24, number of training epochs 300, and the Adam optimizer for optimizing the parameters of the policy network. We also employ a learning rate scheduler that enables the learning rate to decay after every 20 episodes. We found the value of $\lambda = 0.1$ works best across different experimental settings including small and large area search. All our code and models will be made publicly available.

\subsection{AGS Pseudocode}
We present the pseudocode of our proposed AGS algorithm for small area search in algorithm~\ref{A:MASHA}.

\begin{algorithm}
\scriptsize
\caption{The proposed \textsc{AGS} algorithm for training.}
\begin{algorithmic}[1]
\scriptsize
\Require  A search task instance $(x = [x_1,\ldots, x_K]; y = [y_{1}, \ldots y_{r}, \ldots,  y_{K}])$; budget constraint $\mathcal{C}$; Prediction function ($f$) with current parameters $\phi_i^{0} = \phi_i$, i.e., $f_{\phi_i^{0}}$; Search policy ($g$) with current parameters $\zeta_i$, i.e., $g_{\zeta_i}$.; 
\STATE \textbf{Initialize} $o^{0} = [ 0...0]$ ; $B^{t} = \mathcal{C}$; step $t = 0$
\While {$B^{t} > 0 $}
    \State $\tilde{p} = f_{\phi_i^{t}}(x, o^t)$
    \State \emph{j} $\xleftarrow{} \mathit{Sample}_{j \in \{\mathit{1, \ldots i, \ldots, K}\}} [g_{\zeta_i}(\tilde{p}, o^t, B^{t})]$
    \State Query parcel with index $j$ and observe true label $y_{j}$.
    \State Update $\phi_i^{t}$ to $\phi_i^{t+1}$ using $\mathcal{L}_{\mathit{BCE}}$ loss between $\tilde{p}$ and pseudo label $\hat{y}$, defined as   
    $\hat{y}\xleftarrow{}\left\{
     \begin{array}
     {r@{\quad}l}
          \text{$y_{j}$} & \text{if   $y_{j}$ is Observed} \\
          \text{$\tilde{p}_{j}$}  & \text{if  $y_{j}$ is Unobserved.} 
     \end{array}
     \right.$
    \State Obtain reward $R^t = y_{j}$, Update $o^{t}$ to $o^{t+1}$ with $o_{j} = 2y_{j} -1$, and update $B^{t}$ to $B^{t+1}$ with $B^{t+1} = B^{t} - c(k,j)$ (assuming we query $k$'th parcel at $(t-1)$, $c(k,j)$ denotes query cost).
    \State Collect transition tuple $\tau$ at step t, i.e., $\tau^{t}$ = $\bigl(\text{state} = (x, o^t, B^{t})$, action = $j$, reward = $R^{t}$, next state = $(x, o^{t+1} ,B^{t+1})\bigr)$.
    \State \emph{t} $\xleftarrow{} t+1$
\EndWhile \\
\STATE Update search policy parameters $\zeta_i$ using ($\mathcal{L}_{\mathit{RL}}$) based  on the collected transition tuples ($\tau^{t}$) throughout the episode and update initial prediction function parameters $\phi_i$ using ($\mathcal{L}_{\mathit{BCE}}$) based on the collected labels ($y_{j}$) throughout the episode.\\
\STATE \textbf{Return} updated prediction and search policy parameters, i.e., $\phi_{i+1}$ and $\zeta_{i+1}$ respectively.
\end{algorithmic}
\label{A:MASHA}
\end{algorithm}

\subsection{Performance Analysis of different Search Methods across different set of 16K Parcels}

\begin{figure*}
  \centering
  \includegraphics[width=0.85\linewidth]{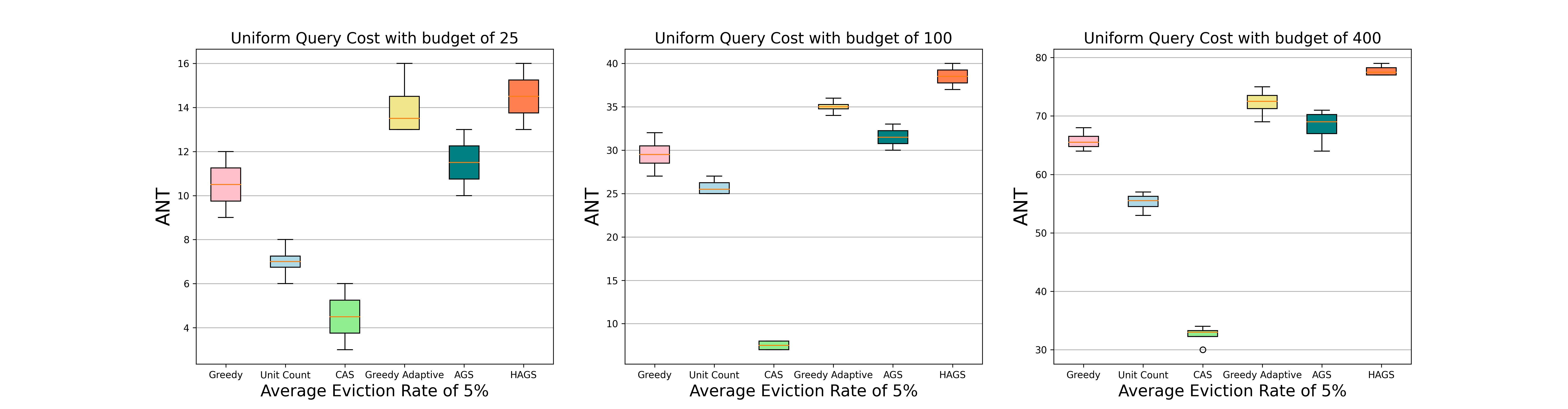}
  \caption{Performance Analysis Across Different Evaluation Set.}
  \label{fig:var_uni_5}
\end{figure*}

\begin{figure*}
  \centering
  \includegraphics[width=0.85\linewidth]{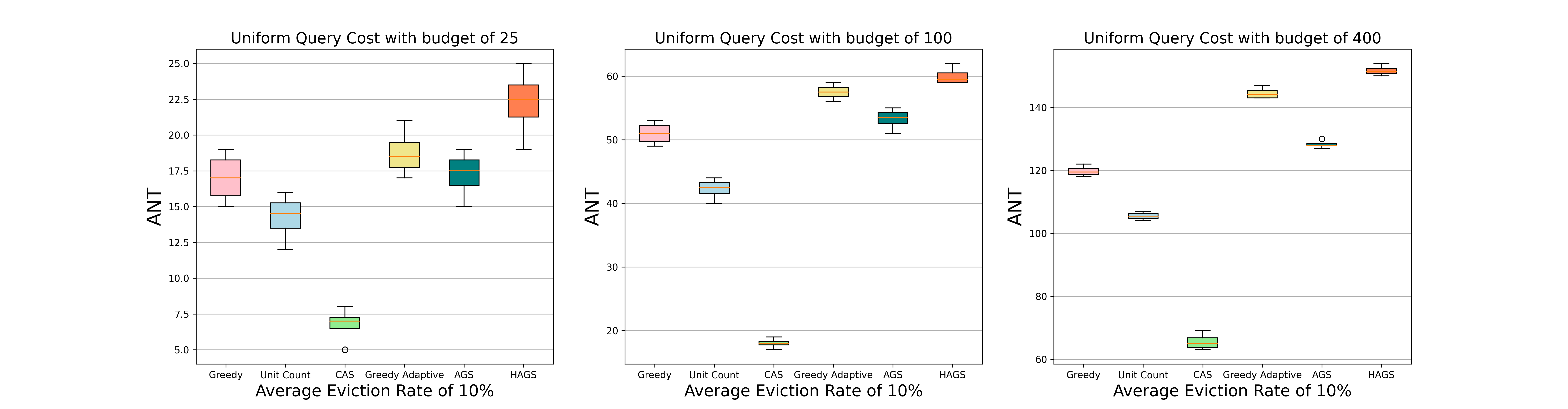}
  \caption{Performance Analysis Across Different Evaluation Set.}
  \label{fig:var_uni_10}
\end{figure*}

\begin{figure*}
  \centering
  \includegraphics[width=0.85\linewidth]{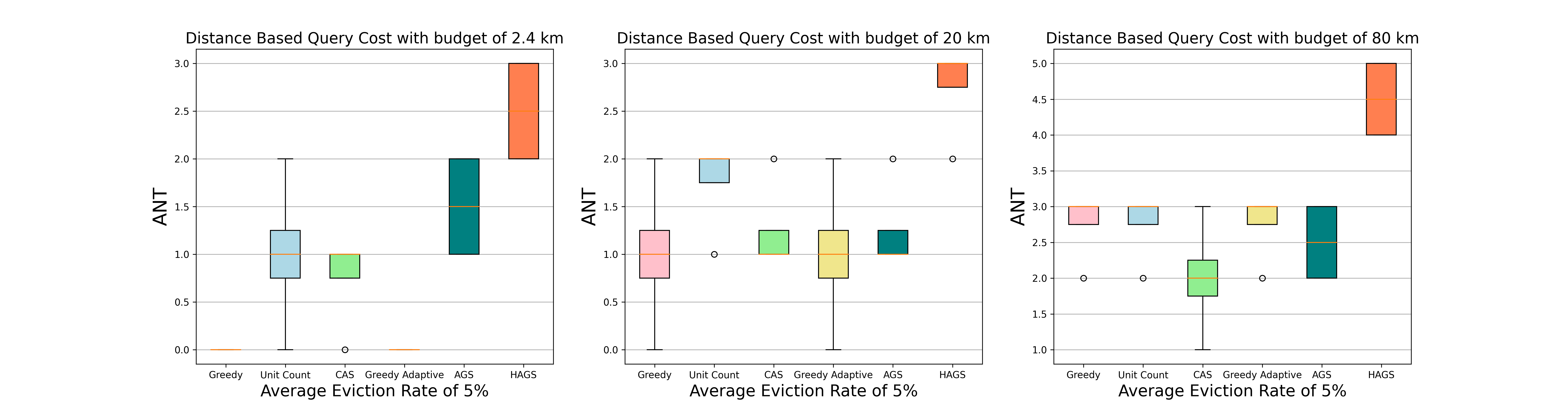}
  \caption{Performance Analysis Across Different Evaluation Set.}
  \label{fig:var_dist_5}
\end{figure*}

\begin{figure*}
  \centering
  \includegraphics[width=0.85\linewidth]{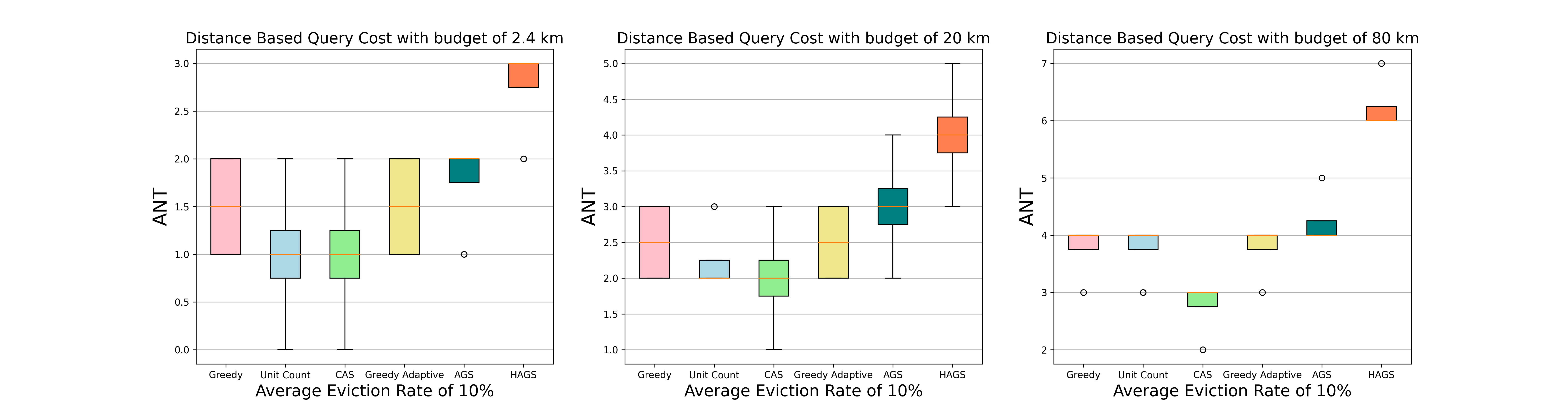}
  \caption{Performance Analysis Across Different Evaluation Set}
  \label{fig:var_dist_10}
\end{figure*}

For comparison, we evaluate the search performance of different methods across 4 different sets of 16k parcels. We report the performance of different methods in terms of ANT in the main paper. Here we present the variance in the search performance of these methods across the same 4 different sets of 16k parcels. In figure~\ref{fig:var_uni_5} and \ref{fig:var_uni_10}, we report the box plot under uniform query cost setting with average eviction rate of 5\% and 10\% respectively. Similarly, in figure~\ref{fig:var_dist_5} and \ref{fig:var_dist_10}, we report the box plot under distance based query cost setting with average eviction rates of 5\% and 10\% respectively.

\begin{figure}
  \centering
  \includegraphics[width=0.95\linewidth]{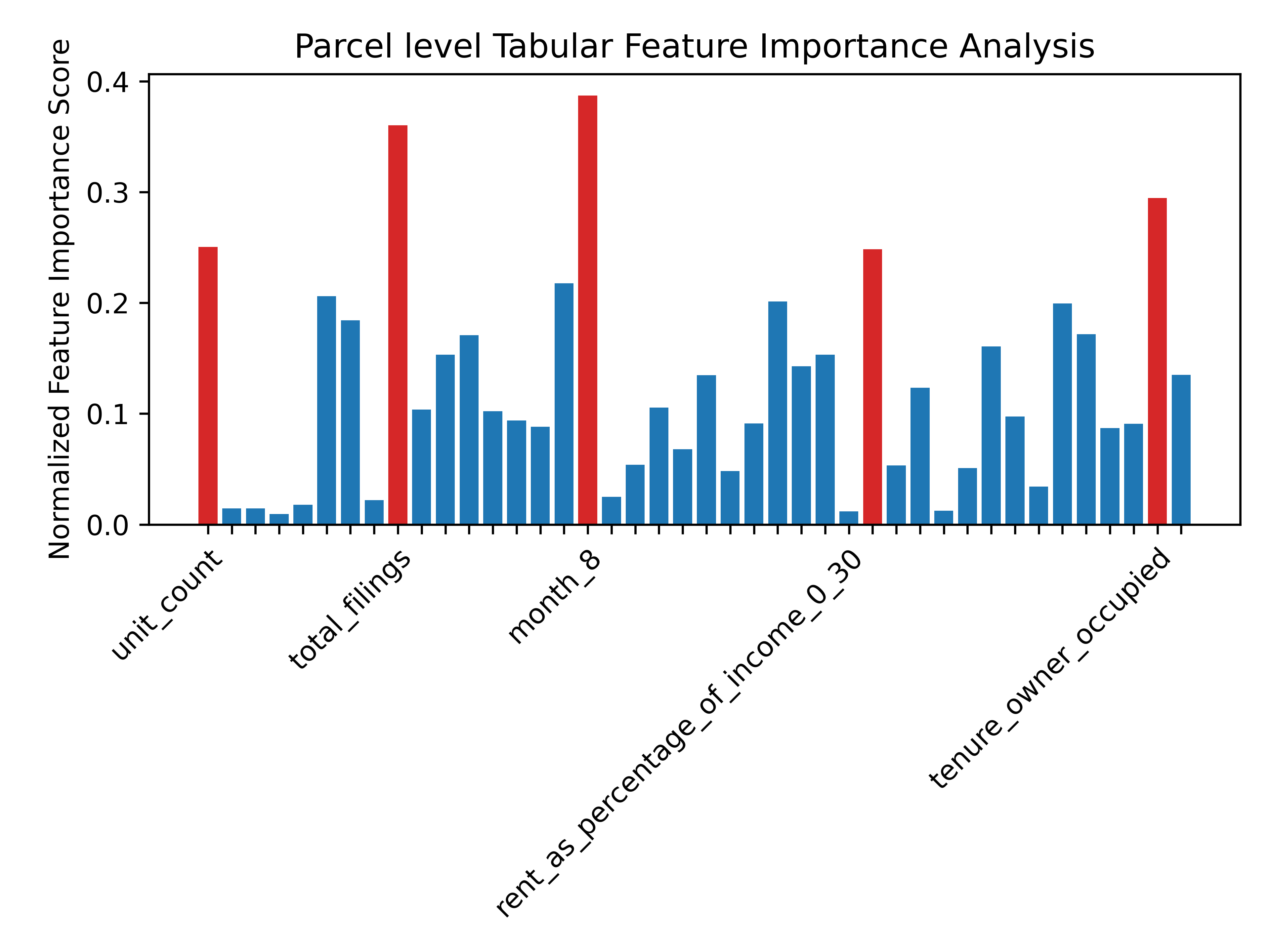}
  \caption{Parcel Level Tabular Feature Importance Analysis}
  \label{fig:tab_feat_imp}
\end{figure}

\subsection{Model Performance by Property Unit Count}
\vspace{3pt}
In the model, the number of units in a property is both explicitly represented as a feature and implicitly represented through satellite imagery or the count of eviction filings in time windows. Since the target is constructed to be binary, that is, whether there will be \emph{any} eviction filing at a property during the eviction period, it is possible that the predictive model is exploiting unit counts to perform the search.

In order to explore this question, the original 26,700 properties are divided into three groups of properties based on unit counts. The first group consists of the smallest properties in the search, those with just two units, of which there are 16,529. The second group has 7,553 properties which all have between three and six units. Finally, we consider the 2,688 properties with over six units. Models are trained and tested using search tasks constructed only from their unit group, with an average positive rate of 10\%.

The results of this analysis are shown in \ref{tab:ant_unit_counts}. Interestingly, searches on the group of properties with between three and six units perform the worst, while more targets are found on average in the two-unit property search. Despite this, both of these searches do not perform as well as the overall search of all properties. When looking at the search of properties with seven units or more, however, the performance observed exceeds that of the combined unit task for search budgets of 20 and 25. These findings raise the question, among others, of whether implementations of Geospatial Active Search in this domain should utilize a single predictive model or a hybrid approach should be employed, with properties stratified by features such as, but not limited to, unit count.

\begin{table*}
    \centering
    \footnotesize
    \caption{\small{Average Number of Targets (ANT) Found by Search Budget and Solution Method for Properties Grouped by Unit Count}}
    \begin{tabular}{p{2.5cm} *{12}{p{0.75cm}}}
        \toprule
        \multicolumn{1}{c}{} & \multicolumn{3}{c}{All} & \multicolumn{3}{c}{2 Units} & \multicolumn{3}{c}{3 to 6 Units} & \multicolumn{3}{c}{7 Units and Above} \\
        \midrule
        Search Budget & $15$ & $20$ & $25$ & $15$ & $20$ & $25$ & $15$ & $20$ & $25$ & $15$ & $20$ & $25$ \\
        \cmidrule(r){1-1} \cmidrule(r){2-4} \cmidrule(r){5-7} \cmidrule(l){8-10} \cmidrule(l){11-13}
        Random & 1.428 & 1.940 & 2.428 & 0.911 & 1.646 & 1.817 & 0.712 & 1.284 & 1.424 & 1.262 & 1.998 & 2.633 \\
        Greedy & 3.796 & 4.296 & 4.780 & 2.825 & 3.968 & 4.201 & 2.204 & 3.251 & 3.602 & 3.659 & 4.471 & 4.912\\ 
        Greedy Adaptive & 3.872 & 4.452 & 5.020 & 3.014 & 4.256 & 4.453 & 2.653 & 3.638 & 4.002 & 3.717 & 5.532 & 5.527 \\ 
        \textbf{AGS} & \textbf{4.372} & \textbf{5.204} & \textbf{5.632} & \textbf{3.524} & \textbf{4.472} & \textbf{5.004} & \textbf{3.156} & \textbf{3.912} & \textbf{4.396} & \textbf{4.280} & \textbf{5.300} & \textbf{5.976} \\
        \bottomrule
    \end{tabular}
    \label{tab:ant_unit_counts}
\end{table*}

\subsection{Parcel Level Tabular Feature Importance Analysis}

In this study, we analyze the importance of 42 distinct parcel level tabular features for active geospatial search using a pre-trained AGS policy network. For this analysis, we calculate the importance score of input features at each step by computing the gradient of the policy network output with respect to the 42 tabular input features. The final step involves averaging these importance scores across all time steps for each input feature and normalizing each value to a range of 0 to 1. The normalized relative importance scores for all 42 features are then presented in figure~\ref{fig:tab_feat_imp}. Note that, in figure~\ref{fig:tab_feat_imp}, we emphasize features (in red) with an importance score exceeding 0.25. We also observe that the following 5 input features are most important to the policy network for active geospatial search: $(1)$ "Unit-count" - encodes the number of units a parcel contains; $(2)$ "total-filings" - encodes the number of total eviction case filings; (3) "month-8" - encodes the number of evictions during the 8'th month; (4) "rent-as-percentage-of-income" - encodes the rent as a percentage of tenant total earnings; and (5) "tenure-owner-occupied" - encodes the time period an owner occupied a particular parcel.

\subsection{Compute Resources}
We use a single NVidia A100 GPU server with a memory of 80 GB for
training and a single NVidia V100 GPU server with a memory of 32 GB for running the inference.

\section{Societal Impact}
Our approach broadly falls into the general category of AI methods that target treatments and interventions in public health, and aims to provide important resources for individuals who are housing insecure, generally in low-resource communities.
Nonetheless, we acknowledge that any future outreach performed using the proposed methods must be carefully considered in close collaboration with domain experts.
One concern is that these methods may enable construction of sensitive datasets that can cause harm to the individuals surveyed if they fall into the wrong hands (for example, are exposed to malicious landlords).  
Consequently, utmost care must be taken to protect these datasets from misuse, such as storing them in secure locations.
Another concern is that our techniques can be used by malicious actors to target disadvantaged individuals and groups.
Indeed, this concern is common across a broad array of AI technologies.
An important redeeming factor in our domain is that our methods require direct cooperation of the individuals being surveyed; it would not be effective without it.
Such cooperation would likely be much easier to obtain for our target application, when domain experts are involved who already have established relationships with the communities being surveyed, but would be difficult to obtain if goals are not benign.